\documentclass[conference]{IEEEtran}
\usepackage{cite}

\ifCLASSINFOpdf
   \usepackage[pdftex]{graphicx}
\else
\fi
\usepackage{amsmath}
\usepackage{verbatim}
\begin{document}
%
\title{Text Line Segmentation for Challenging Handwritten Document Images Using Fully Convolutional Network}

\author{\IEEEauthorblockN{}}


%
\author{\IEEEauthorblockN{Berat Barakat,
Ahmad Droby,
Majeed Kassis and
Jihad El-Sana}
\IEEEauthorblockA{The Department of Computer Science\\
Ben-Gurion University of the Negev\\
Email: \{berat, drobya, majeek, el-sana\}@cs.bgu.ac.il}
}


\maketitle

\begin{abstract}
This paper presents a method for text line segmentation of challenging historical manuscript images. These manuscript images contain narrow interline spaces with touching components, interpenetrating vowel signs and inconsistent font types and sizes. In addition, they contain curved, multi-skewed and multi-directed side note lines  within a complex page layout. Therefore, bounding polygon labeling would be very difficult and time consuming. Instead we rely on line masks that connect the components on the same text line. Then these line masks are predicted using a Fully Convolutional Network (FCN). In the literature, FCN has been successfully used for text line segmentation of regular handwritten document images. The present paper shows that FCN is useful with challenging manuscript images as well. Using a new evaluation metric that is sensitive to over segmentation as well as under segmentation, testing results on a publicly available challenging handwritten dataset are comparable with the results of a previous work on the same dataset.
\end{abstract}

%
\IEEEpeerreviewmaketitle

\section{Introduction}

\begin{figure}
\centering
\includegraphics[width=3.1in]{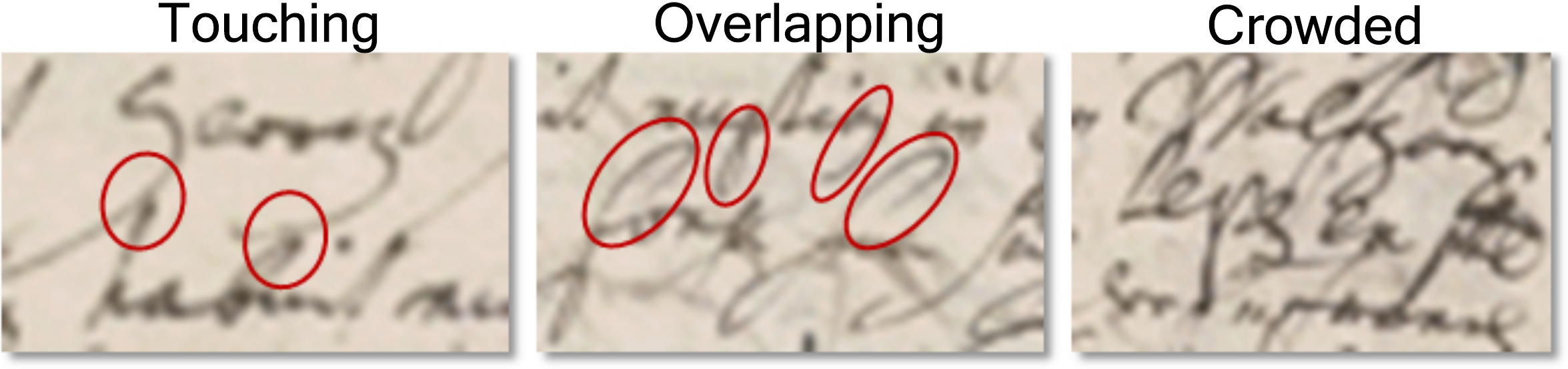}
\caption{Text line segmentation problems with regular handwritten documents}
\label{handwritten}
\end{figure}

Historical handwritten documents are valuable since they connect past and present. Commonly they are converted into digital form for being easily available to scholars worldwide. However, digital historical documents pose real challenges for automatic writer identification, keyword searching, and indexing. Text line segmentation of document images is an essential pre-processing operation for these automatizing problems. The problems for text line segmentation of handwritten documents consist of touching, overlapping and crowded characters and vowel signs among consecutive text lines besides narrow interline spacing (Figure~\ref{handwritten}).

\begin{figure}
\centering
\includegraphics[width=3.1in]{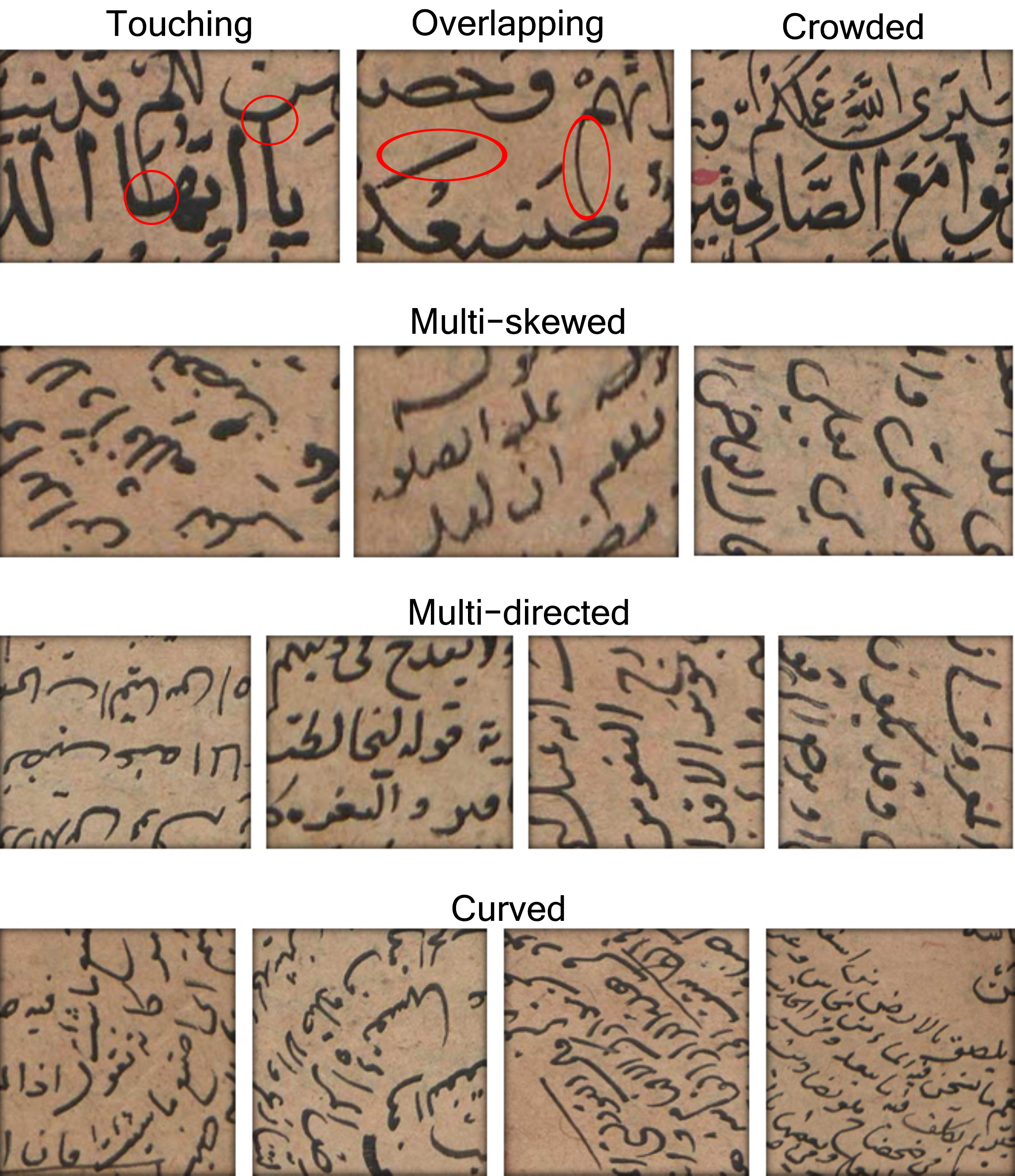}
\caption{Additional text line segmentation problems with challenging handwritten documents. Various writing styles are also noticeable.}
\label{challenging}
\end{figure}

In addition to the problems of handwritten documents, challenging handwritten documents contain various writing styles with inconsistent font types and  font sizes through multi-skewed, multi-directed and curved text lines (Figure~\ref{challenging}).

Several text line extraction methods for handwritten documents have been proposed. Projection method was initially used for printed documents \cite{ha1995document,manmatha1999scale} then modified for skewed \cite{arivazhagan2007statistical,bar2009line} and multi-skewed documents \cite{ouwayed2012general}. Smearing method \cite{wong1982document} which fills the space between consecutive foreground pixels can be used on skewed documents \cite{alaei2011new} as well. Grouping method aggregates pixels or connected components in a bottom up strategy and is superior in case of skewed and curved text lines \cite{bukhari2009script,cohen2014using}. Machine learning algorithms, a type of grouping method, handle text line segmentation as a pixel classification problem. Pixel classification can be done in a sliding window manner \cite{moysset2015paragraph, pastor2016complete} which is not desirable due to redundant and expensive computation of overlapping areas in the sliding windows. On the other hand, dense prediction does not suffer from redundant computation and has been successfully used for text line segmentation of handwritten documents \cite{vo2016dense,renton2017handwritten}.

However, text line extraction of challenging documents has not been extensively studied. This paper provides a dataset (https://www.cs.bgu.ac.il/~vml/) of challenging documents with multi-skewed, multi-directed and curved handwritten text lines. It then describes text line segmentation of this dataset using Fully Convolutional Network (FCN). We also propose a new evaluation metric that is sensitive to both, over and under segmentation of lines in contrast to the available metrics. Using the new evaluation metric we show that FCN based method is comparable to Cohen et al.'s method \cite{cohen2014using}.

In the rest of the paper we describe our method and the new evaluation metric in detail, and present the challenging dataset and report experimental results. After comparing results we conclude and discuss the future directions.

\section{Method}
\label{sec:Method}

Fully Convolutional Network has made great improvements in object segmentation field \cite{long2015fully}. It is an end to end semantic segmentation framework that extracts the features and learns the classifier function simultaneously. FCN inputs the original images and their pixel level annotations for learning the hypothesis function that can predict whether a pixel belongs to a text line label or not. So the crucial question is how to annotate the text lines. Baseline labeling is not applicable to all the alphabets whereas bounding polygon is applicable but very cumbersome for crowded documents \cite{gruning2017read}. Instead of baseline or bounding polygon, we used line mask labeling that connects the characters in the same line (Figure~\ref{labeling}). A line mask disregards diacritics and touching components between lines.

\subsection{FCN architecture}
\begin{figure*}[!t]
\centering
\includegraphics[width=6in]{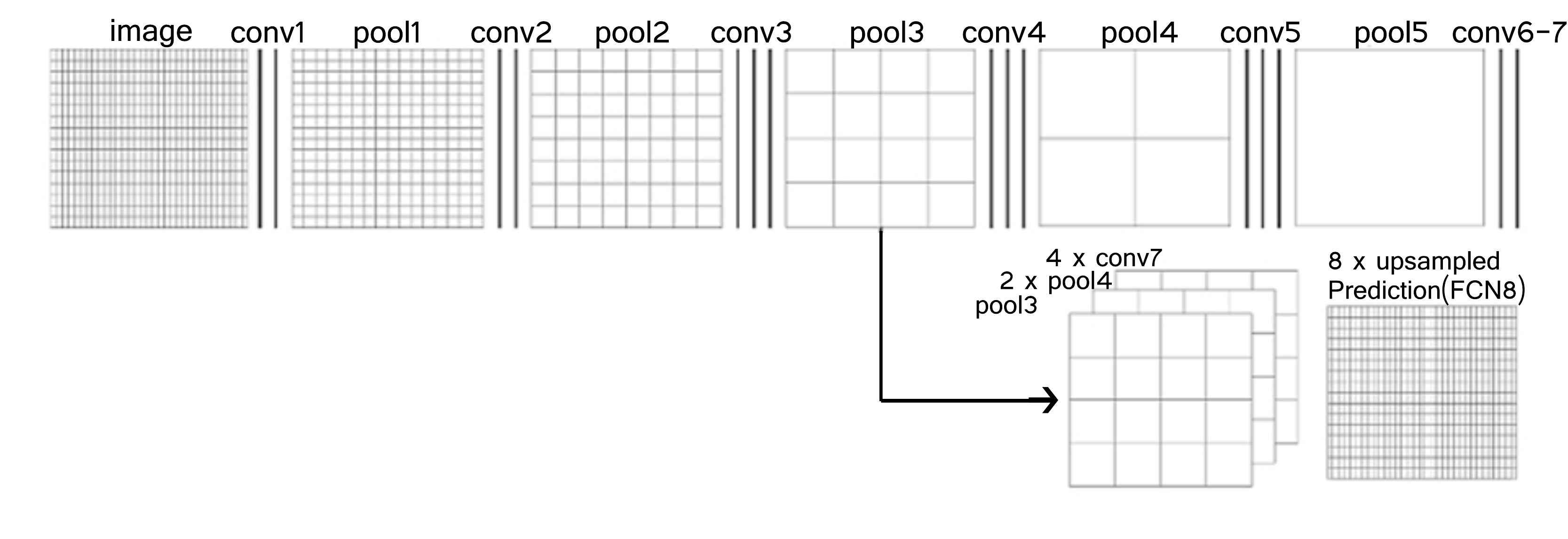}
\caption{The FCN architecture. Pooling and prediction layers are shown as grids that show relative coarseness. Convolutional layers are shown as vertical lines. FCN8 4 times upsamples the final layer, 2 times upsamples the pool4 layer and combine them with pool3 layer finally to upsample to input size.}
\label{arch}
\end{figure*}

The FCN architecture (Figure~\ref{arch}) we used is based on the FCN proposed for semantic segmentation \cite{long2015fully}. First five blocks, encoder part, follow the design of VGG 16-layer network \cite{simonyan2014very} except the discarded final layer. The encoder consists of five convolutional blocks. Each convolutional block contains a number of convolutional layers followed by a max pooling layer. Through the encoder, input image is downsampled, and the filters can see coarser information with larger receptive field. Then the decoder part of FCN upsamples coarse outputs to dense pixels. Upsampling is done by transpose convolution by applying a convolution filter with a stride equal to $\frac{1}{f}$, for upsampling by a factor $f$.

FCN has two types, FCN8 and FCN32, according to the upsampling factor used in the last layer. FCN32 upsamples the last convolutional layer by $f=32$ at one time. However, particularly FCN8 architecture was selected because it has been successful in page layout analysis of a similar dataset \cite{kurar2018binarization}. FCN8 adds final layer of encoder to the lower layers with finer information, then upsamples the combined layer back to the input size. Default input size of VGG is $224\times224$, which does not cover more than 2 main text lines and 3 side text lines. To include more context we changed the input size to $320\times320$ pixels. We also changed the output channel to 2 which is the number of classes, text line or background.

\subsection{Pre-processing}
\begin{figure}
\centering
\includegraphics[width=3.4in]{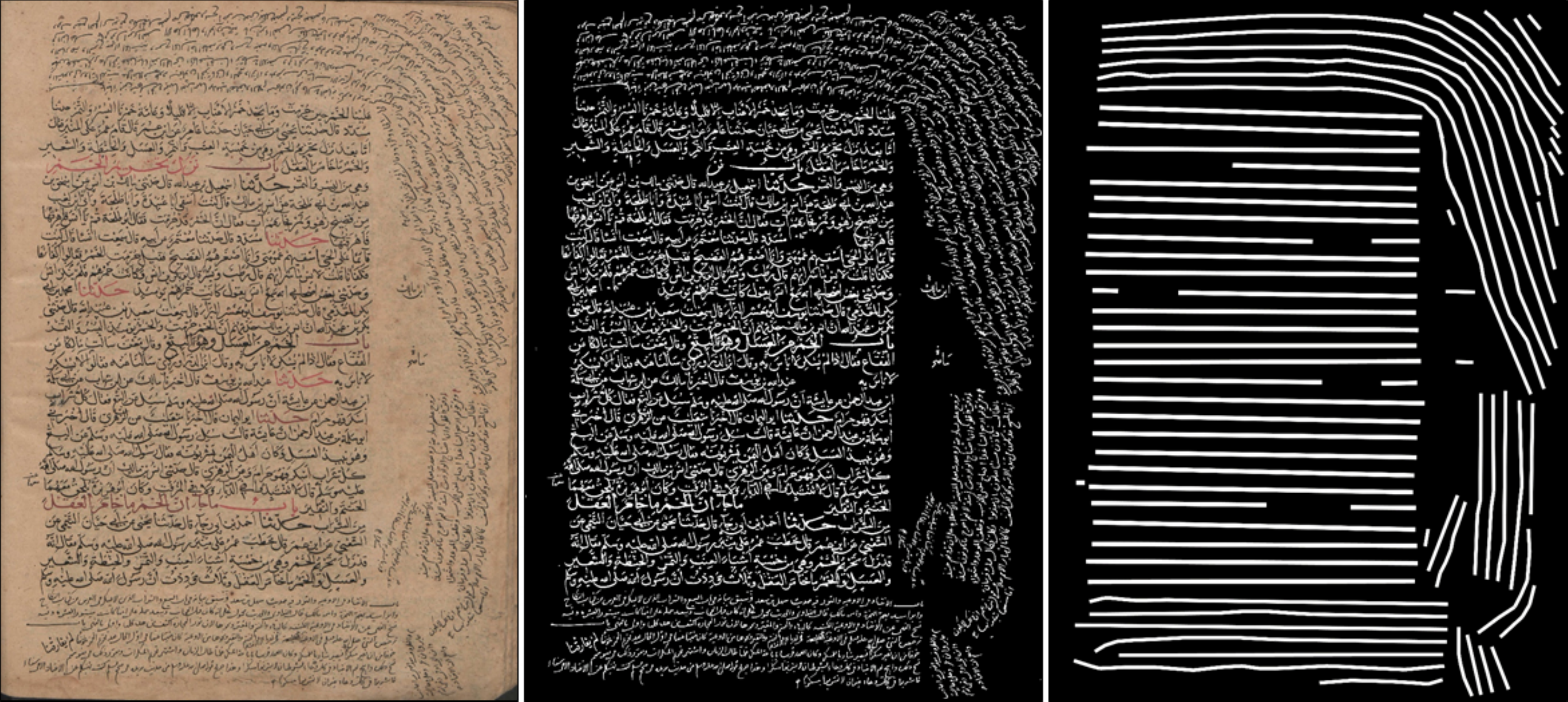}
\caption{A sequence of original, binarized and labeled document images. Random patches for training are generated from the binarized and labeled images.}
\label{labeling}
\end{figure}

We binarize the 30 document images, each with an approximate size of $3000 \times 4000$, by applying an adaptive binarization method for historical documents \cite{bar2007binarization}. Binarized images were inverted before inputting them to the FCN. Then we manually annotated the line masks on the document images. A sequence of original, binarized and labeled document images is demonstrated in Figure~\ref{labeling}. Finally a total of $50.000$ and $6.000$ random patches of size $320\times320$ were generated for training and validation sets of each fold respectively.

\subsection{Training and testing}
We applied 6 fold cross validation scheme for the experiments. Each fold was split into train, validation and test sets. In each fold, training continued for 80 epochs and the model with the least validation loss value was saved. The best model was then used to predict the corresponding test set. This training procedure ensures generalizability of the proposed model. The FCN was trained by a batch size of 16, using Stochastic Gradient Descent (SGD) with momentum equals to $0.9$ and learning rate equals to $0.001$. VGG was initialized with its publicly available pre-trained weights.

During the testing, a sliding window of size $320\times320$ was used for prediction, but only the inner window of size $100\times100$ was considered. Page was padded with black pixels at its right and bottom sides if its size is not an integer multiple of the sliding window size, in addition to padding it at 4 sides for considering only the central part of the sliding window.

\subsection{Post-processing}
Occasionally predicted line masks were disconnected. Thus, we needed to post-process the FCN output. Given a predicted line mask image, firstly the orientation of each connected component was computed. Then the image was split into $N$ layers where each layer contains the connected components with same orientation. Later a directional morphological operation was applied on each layer. Resulting layers at the end were combined back using a pixel-wise OR operation.

Let $C=\{c_1,c_2, ..., c_M\}$ is the set of connected components in the predicted line mask image. $C$ is further divided into $N$ intersecting subsets $B_1, B_2, ..., B_N \subseteq C$ such that:
\begin{equation}
B_i = \{c_i: \alpha(c_i)^2|v_j^T \cdot \theta(c_i)| < \epsilon \}
\end{equation}
$$i=1,2,\dots M, j=1,2,\dots N$$

\begin{equation}
v_j = (\cos(j \frac{\pi}{N}), \sin(j \frac{\pi}{N}))
\end{equation}
\begin{equation}
\alpha(c)= \frac{R_{maj}}{R_{maj} + R_{min}}
\end{equation}
where $v_j\in[0,\pi]$ is a particular orientation and $\epsilon\in[0,1]$ is the threshold for selecting the connected components perpendicular to this particular orientation. $R_{maj}$ and $ R_{min}$ are the major and minor axes of the fitted ellipse to the connected component $c$ respectively. $\alpha(c)\in[0.5,1]$ indicates how sure are we about the orientation of the component $c$.
$\theta(c)$ is the unit vector that represents the orientation of the fitted ellipse to the connected component $c$. Ellipse fitting was done using the algorithm described in \cite{fitzgibbon1996buyer}.

\begin{figure}
\centering
\includegraphics[width=3.4in]{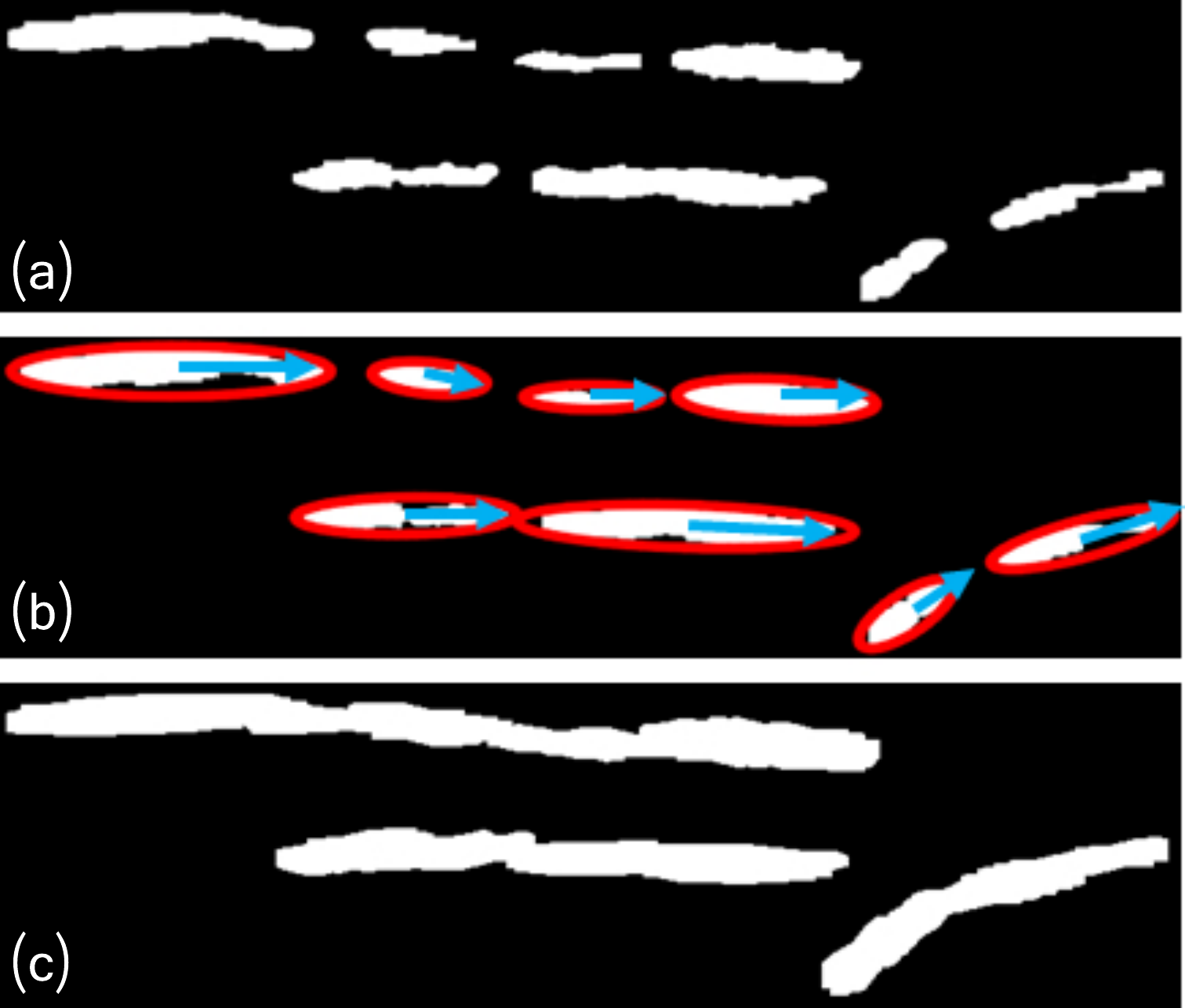}
\caption{Post processing phases: (a) Predicted line mask may have disconnected components. (b) For each component an ellipse (red) is fitted and its orientation vector $\theta(c)$ (blue) is computed. (c) Morphological dilation is applied to each component with a narrow kernel in the direction of its fitted ellipse. }
\label{post}
\end{figure}

Eventually for each subset $B_i$ a morphological operation with a narrow kernel in the orientation of this subset was applied. Figure~\ref{post} shows the result of post-processing on a sample predicted line mask image. 

\subsection{Connectivity Component Based Line Extraction Accuracy Metric}
Available evaluation methods for text line segmentation either use a pixel-wise matching mostly normalized by line length or maximum overlap according to a certain threshold between the extracted and annotated lines. These methods have their short-comings. Thus, we present a different evaluation method that provides a better picture of the results. 

The theoretical basis is as follows. A line extraction algorithm succeeds in extracting a complete text line if it has succeeded in finding all the connected components of this line. That is if the algorithm labels all the connected components of a line with the same label, then it has successfully extracted this line without any errors. This is in contrast to having multiple labels, over segmentation, or extracting part of the connected components, under segmentation, along the same text line.

To describe the new metric, we define the term {\em connectivity component}. A connectivity component is the connection between two consecutive components with the same label. The number of connectivity components in a line is equal to the number of connectivity components between every two consequent connected components and in addition to it a beginning of line connectivity component. The extra connectivity component handles cases where a line contains one connected component only. {\emph Complete extraction of a line} with several connectivity components is extracting all its connectivity components and assigning them the same label.

To quantify the new metric we define recall and precision for calculating F-measure. Recall is the number of connectivity components extracted by the algorithm in a line, out of all connectivity components found in the corresponding line in ground truth. Precision is the number of correct connectivity components extracted by the algorithm in a line out of all connectivity components extracted by the algorithm. Note that some connectivity components extracted by the algorithm are not found in the ground truth, and some connectivity components are found in the ground truth but not extracted by the algorithm. First type of error is quantified in the precision part of the metric, while the latter type of error is quantified in the recall part of the metric.

\begin{figure}
\centering
\includegraphics[width=3.3in]{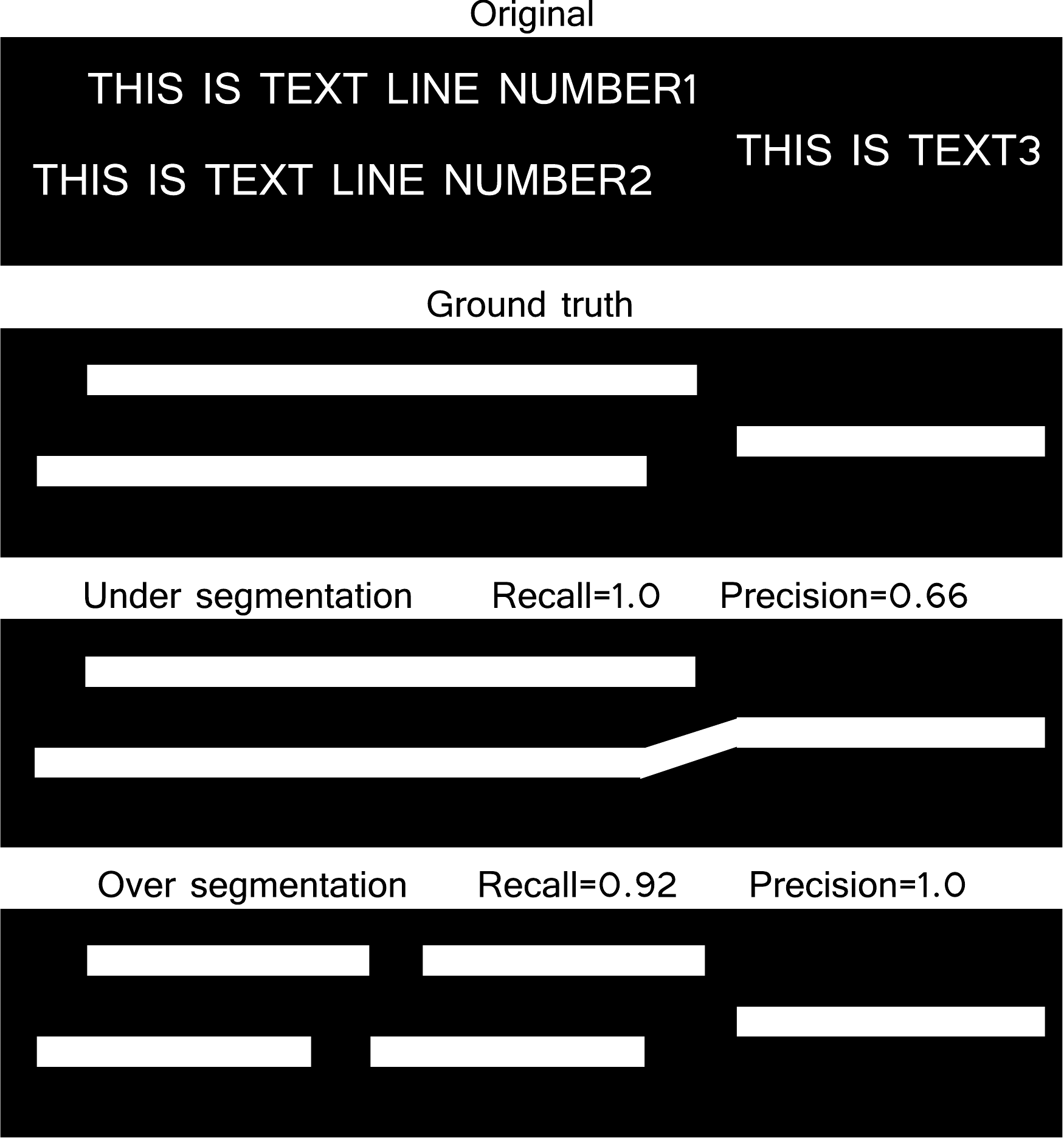}
\caption{Connectivity component based metric penalizes under segmentation by its precision definition and over segmentation by its recall definition.}
\label{metric}
\end{figure}

Let $G=\{g_1,g_2,g_3,\dots g_m\}$ is the set of connected components of a line in the ground truth, $E_i\in\{E_1,E_2,E_3,\dots E_n\}$ is the set of extracted lines such that $E_i\cap G \neq \emptyset$,
then for this line in the ground truth, recall ($R$) and precision ($P$) is:
\begin{equation}
R=\sum\limits_{i}{\frac{|E_i\cap G|-1}{|G|-1}}
\label{eg:r}
\end{equation}
\begin{equation}
P=\frac{\sum\limits_{i}{|E_i\cap G|-1}}{\sum\limits_{i}{|E_i|-1}}
\label{eq:p}
\end{equation}
The recall definition penalizes over segmentation of a line where an extraction algorithm assigns multiple labels to the components of a single line. In contrast, the precision definition penalizes under segmentation where an extraction algorithm incorrectly assigns a line label to the components that are not in the ground truth of this line (Figure~\ref{metric}).

\section{Dataset}
Although several benchmark datasets \cite{gatos2011icdar2009,stamatopoulos2013icdar,diem2017cbad} of handwritten document images are available, a challenging document dataset is absent. We collected a set of challenging document images from the Islamic Heritage Project (IHP), Harvard. This dataset is publicly available (https://www.cs.bgu.ac.il/~vml/).

The challenging dataset contains 30 pages from two different manuscripts. It is written in Arabic language and contains 2732 text lines where a considerable amount of them are multi-directed, multi-skewed or curved. Ground truth where text lines were labeled manually by line masks is also available in the dataset.

\section{Results}

We tested the proposed system on the new challenging handwritten document dataset. In each fold we trained FCN on 50.000 patches randomly cropped from 20 pages, validated on 6.000 patches randomly cropped from 5 pages and tested on 5 whole pages using a sliding window. Predicted line mask images were then post-processed with $N=10$ and $\epsilon=0.2$. Extracted text lines were evaluated using the new metric to calculate the F-measure.

\begin{figure}
\centering
\includegraphics[width=3.2in]{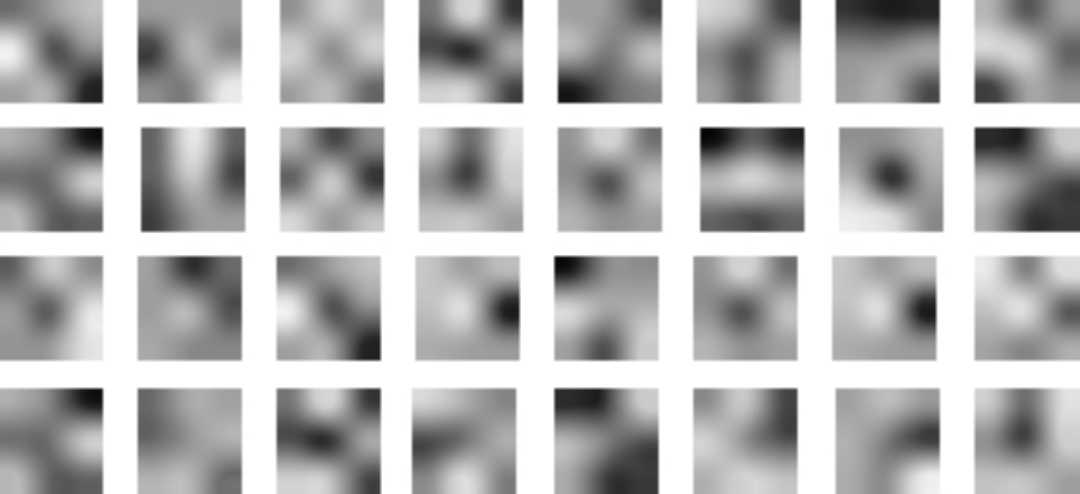}
\caption{Visualization of the filters in the first convolutional layer.}
\label{filters}
\end{figure}

Entire training took around 9 days. Visualization of the first convolutional layer filters shows that network have learned and filters have converged (Figure~\ref{filters}). The model achieved $89\%$ training accuracy and $88\%$ validation accuracy on average. Two characteristics of the dataset lead the model lacking to overfit to the training set. First it contains two manuscripts with 6 and 24 pages. The manuscript with 6 pages caused most of the errors. Second, although dataset contains considerable amount of  multi-skewed, multi-directed and curved lines, they spatially cover smaller area due to smaller font size. This lead to less number of random patches with skewed or curved lines in relative to the number of random patches with regular lines.

Table~\ref{result} shows the performance of our method compared with the method of Cohen et al.\cite{cohen2014using}. Their approach achieved outstanding results on ICDAR2009 \cite{gatos2011icdar2009} and ICDAR2013 \cite{stamatopoulos2013icdar} datasets. We run their publicly available code (http://www.cs.bgu.ac.il/~rafico/LineExtraction.zip) on the challenging handwritten dataset.

\begin{table}[!t]
\caption{Comparison with the method of Cohen et al.}
\label{result}
\centering
\begin{tabular}{c|c|c|c}
Method                              & Recall    & Precision & F-measure\\
\hline
Proposed                            & \textbf{0.82}      & \textbf{0.78}      & \textbf{0.80 }      \\
Cohen et al.\cite{cohen2014using}   & 0.74      & 0.60      & 0.66        \\
\end{tabular}
\end{table}

\begin{figure*}
\centering
\includegraphics[width=7in]{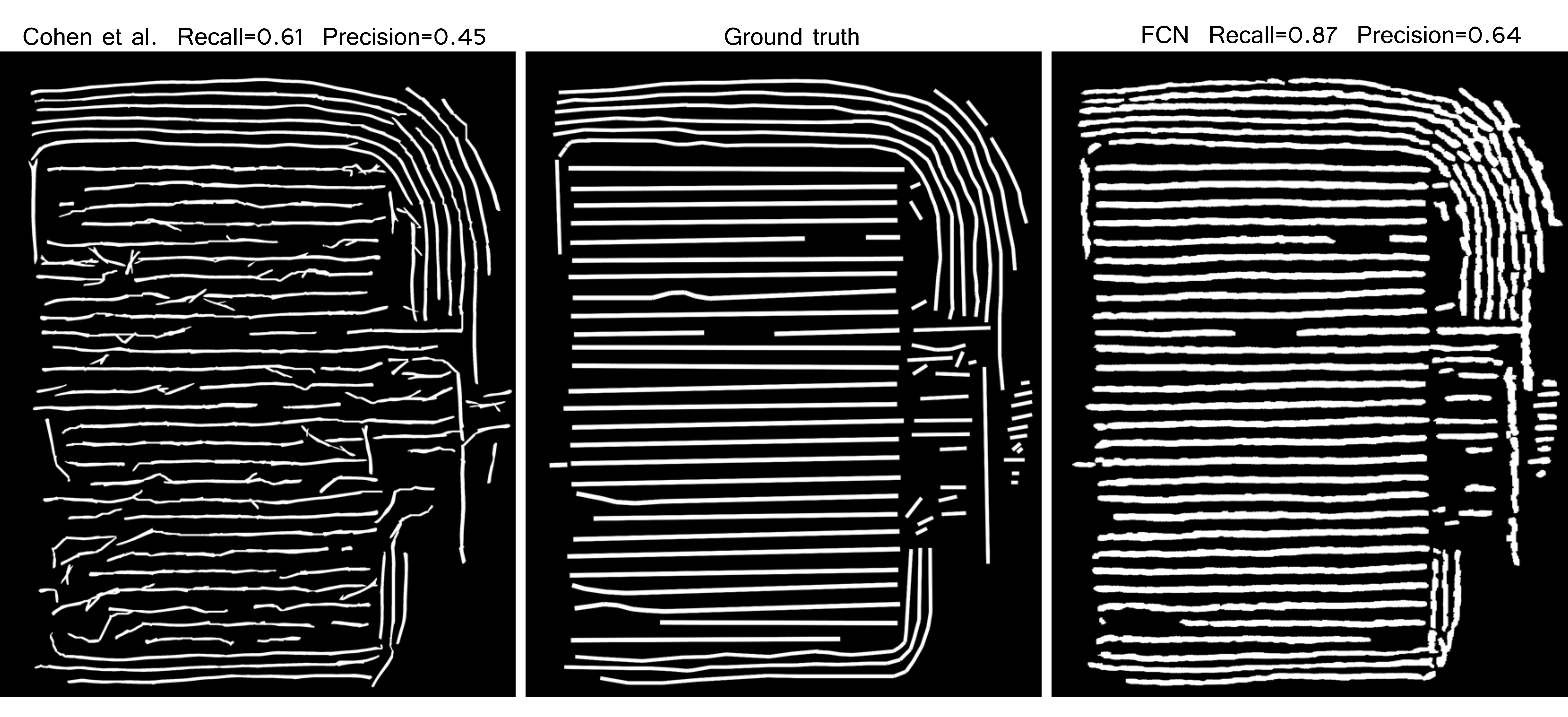}
\caption{Sample image of ground truth and corresponding outputs of Cohen et al. \cite{cohen2014using} and FCN. Lower precision values show that both method tend to under segment. Most errors of FCN method occur at curved areas whereas most errors of method of Cohen et al. occur at the main text areas.}
\label{output}
\end{figure*}

Our method outperforms the method of Cohen et al. in terms of both recall and precision. Both methods have lower precision values than recall values. This demonstrates that most of their errors are due to wrongly connected lines in their output. Therefore both method tend to under segment more than over segment. We have noticed that in the output of our method, wrongly connected lines mostly crop up at the curved areas in contrast to the output of Cohen et al where the wrongly connected lines are mostly crop up at the main text areas. The former was a result of small number of training patches with curved lines. Curved lines can be long but their curved part covers relatively a small spatial area which is one or two corner parts of a page. The latter was a result of small number of main text lines in relative to the number of side text lines in a page, where the average height of text lines converges to the height of side text lines. Therefore method of Cohen et al., which runs according to the average height of text lines, has most errors in main text areas. Figure~\ref{output} shows some qualitative results for the latter and the former types of errors on the challenging dataset.

\section{Conclusion}
This paper introduces challenging handwritten documents, presents a dataset of challenging handwritten documents and its text line segmentation using FCN. Line mask labeling is less cumbersome for challenging handwritten documents and is a proper way for FCN training. We have also defined a new evaluation metric with the concept of connectivity component. This metric is sensitive to both over and under segmentation. New metric is used to validate the proposed method on the challenging handwritten dataset.
As a future work, performance on curved text lines can be improved by augmenting patches with curved lines.

\section*{Acknowledgment}
The authors would like to thank the support of the Frankel Center for Computer Science at Ben-Gurion University of the Negev.



%
\bibliographystyle{IEEEtran}
\bibliography{ref}
%
\end{document}